\documentclass[letterpaper, 10 pt, journal, twoside]{IEEEtran}

\usepackage{amsmath, amssymb, amsfonts} 
\usepackage{mathptmx} 
\usepackage{times} 
\usepackage{bm} 

\usepackage{graphicx} 
\usepackage{epsfig} 
\usepackage{graphics} 
\usepackage{float} 
\usepackage{subfigure} 
\usepackage{threeparttable} 
\usepackage{booktabs} 
\usepackage{multirow} 
\usepackage{caption}
\usepackage{array} 
\usepackage{tabularx} 
\usepackage{colortbl} 
\usepackage[table]{xcolor} 
\definecolor{discblue}{RGB}{235,245,255}
\definecolor{genegreen}{RGB}{235,255,235}
\usepackage{algorithm}
\usepackage{enumitem} 

\usepackage{algorithmic} 
\usepackage[algo2e]{algorithm2e}  

\usepackage{cite} 
\usepackage[colorlinks,linkcolor=blue,citecolor=blue]{hyperref} 
\usepackage[capitalize]{cleveref} 

\usepackage{textcomp}
\usepackage{indentfirst} 
\usepackage{verbatim} 
\usepackage{balance} 
\usepackage[marginal]{footmisc} 

\usepackage{pifont} 
\usepackage[notextcomp]{stix} 
\usepackage{orcidlink} 

\usepackage{soul} 
\sethlcolor{yellow}
\usepackage{placeins}

\soulregister{\cite}7
\soulregister{\cref}7

\def\BibTeX{{\rm B\kern-.05em{\sc i\kern-.025em b}\kern-.08em
    T\kern-.1667em\lower.7ex\hbox{E}\kern-.125emX}}

\title{VistaDepth: Improving far-range Depth Estimation with Spectral Modulation and Adaptive Reweighting}
\author{\IEEEauthorblockN{Anonymous Authors}
}

\author{Mingxia Zhan, Li Zhang, Yingjie Wang, XiaoMeng Chu, Beibei Wang and Yanyong Zhang%
\thanks{Mingxia Zhan is with the Hefei University of Technology, Hefei, Anhui, China (email: mxzhan@mail.hfut.edu.cn)}%
\thanks{Li Zhang is with the Hefei University of Technology, Hefei, Anhui, China (email: lizhang@hfut.edu.cn)}%
\thanks{Yingjie Wang is with the University of Science and Technology of China, Hefei, Anhui, China (email: yingjiewang@mail.ustc.edu.cn)}%
\thanks{Xiaomeng Chu is with the Institute of Artificial Intelligence, Hefei Comprehensive National Science Center, Hefei, Anhui, China (email: cxmeng@mail.ustc.edu.cn)}%
\thanks{Beibei Wang is with the Institute of Artificial Intelligence, Hefei Comprehensive National Science Center, Hefei, Anhui, China (email: wbb@iai.ustc.edu.cn)}%
\thanks{Yanyong Zhang is with the University of Science and Technology of China, Hefei, Anhui, China (email: yanyongz@ustc.edu.cn)}%
}

\begin{document}
\maketitle
\thispagestyle{empty} 
\pagestyle{empty}   

\begin{abstract}
Monocular depth estimation (MDE) aims to infer per-pixel depth from a single RGB image. While diffusion models have advanced MDE with impressive generalization, they often exhibit limitations in accurately reconstructing far-range regions. This difficulty arises from two key challenges. First, the implicit multi-scale processing in standard spatial-domain models can be insufficient for preserving the fine-grained, high-frequency details crucial for distant structures. Second, the intrinsic long-tail distribution of depth data imposes a strong training bias towards more prevalent near-range regions. To address these, we propose VistaDepth, a novel diffusion framework designed for balanced and accurate depth perception. We introduce two key innovations. First, the Latent Frequency Modulation (LFM) module enhances the model's ability to represent high-frequency details. It operates by having a lightweight network predict a dynamic, content-aware spectral filter to refine latent features, thereby improving the reconstruction of distant structures. Second, our BiasMap mechanism introduces an adaptive reweighting of the diffusion loss strategically scaled across diffusion timesteps. It further aligns the supervision with the progressive denoising process, establishing a more consistent learning signal. As a result, it mitigates data bias without sacrificing training stability. Experiments show that VistaDepth achieves state‑of‑the‑art performance for diffusion‑based MDE, particularly excelling in reconstructing detailed and accurate depth in far‑range regions.
\end{abstract}


\section{Introduction}
\label{sec:intro}
MDE, the task of inferring dense depth from a single RGB image \cite{thorough_review}, is a cornerstone of 3D computer vision \cite{wang2023multi}. Recovering 3D geometry from a 2D projection is an inherently ill-posed problem, demanding strong reliance on learned priors such as object geometry, scene context, and occlusion reasoning \cite{depthanythingV2}. Accurate far-range depth perception is critical for applications from autonomous driving \cite{Li2023} to augmented reality \cite{vr}, where reliable understanding of distant geometry underpins safe navigation and interaction.

However, this vital capability is fundamentally undermined by an inherent property of real-world scenes: an uneven depth distribution. As our analysis across multiple depth datasets confirms in Figure \ref{fig:depth_distribution} (a), near-range areas command a high concentration of pixels, while far-range regions exhibit sparse, long-tail characteristics. This disparity imposes a strong training bias towards the data-rich foreground, leading to a persistent and critical failure in accurately resolving far-range structures. Previous MDE algorithms can be broadly categorized into two distinct paradigms: discriminative-learning-based and diffusion-based. The discriminative learning paradigms directly developed targeted strategies to conquer this imbalance. For instance, some approaches employed distance-aware loss functions \cite{jiao2018look} to prioritize underrepresented far-range regions, while others used ensemble architectures \cite{dme} to dedicate specialized components to distinct depth intervals. However, such discriminative models are typically trained on limited datasets and exhibit significant dependence on the underlying data distributions, leading to poor generalization. To overcome the problem of poor generalization, diffusion-based frameworks like Marigold \cite{marigold} marked a significant paradigm shift. By fine-tuning the U-Net of a pre-trained Stable Diffusion backbone \cite{rombach2022high}, these methods successfully addressed the critical issue of generalization, achieving high-quality depth predictions even with limited synthetic data. Despite the progress, the existing methods still struggle with accurate prediction in far-range regions. This persistent challenge stems from two factors unique to this fine-tuning approach. First, the pre-trained backbone exhibits a strong spectral bias~\cite{rahaman2019spectral}, making it inherently difficult to learn high-frequency details from sparse data. Second, the fine-tuning is skewed by a severe supervision bias, as the optimization is dominated by dense, near-range pixels, leading to the degradation of far-range structures~\cite{karras2022edm}.

\begin{figure*}[t]
  \centering
  \includegraphics[width=0.9\textwidth]{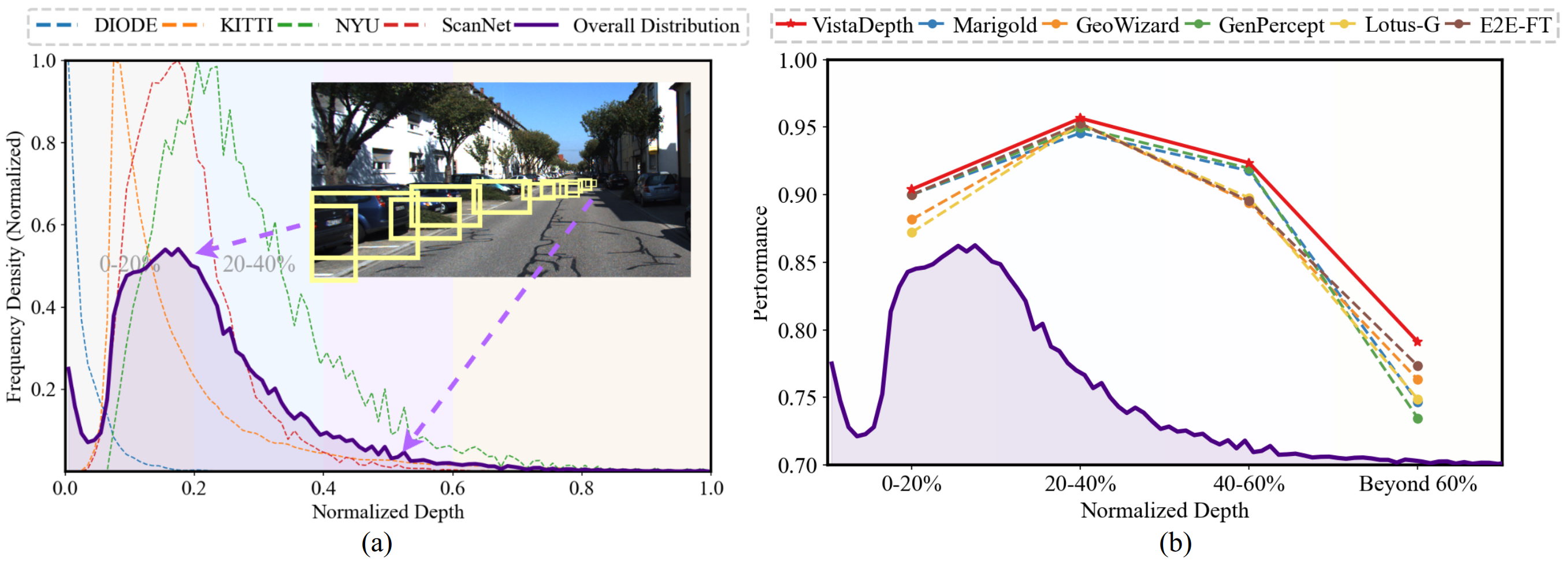}
  \vspace{-0.35cm}
\caption{
\textbf{(a)} Illustration of the long-tail depth distribution in a typical outdoor scene. Near-range pixels are dense, while far-range pixels are sparse.
\textbf{(b)} VistaDepth demonstrates superior performance across all depth ranges, with a disproportionately large gain in the challenging, data-sparse far-range regions, validating its effectiveness in mitigating data bias.}
\label{fig:depth_distribution}
  \label{fig:depth_distribution}
\end{figure*}

To overcome these bottlenecks, we propose VistaDepth, a unified framework that enhances diffusion-based MDE through two complementary refinements. At the feature level, we introduce the LFM module to counteract the inherited spectral bias of the backbone \cite{rahaman2019spectral}. Inspired by recent works that demonstrate the power of explicit frequency-domain processing~\cite{ye2024diffusionedge}, the LFM module is embedded within the U-Net's decoder to provide a dedicated mechanism for dynamic feature refinement. It operates by enhancing critical frequency components within the latent features, thereby improving the reconstruction of distant structures. Simultaneously, at the supervisory level, our BiasMap mechanism introduces an adaptive reweighting of the diffusion loss. The varying signal-to-noise ratio (SNR) across diffusion timesteps makes traditional static loss reweighting ill-suited. Consequently, naively applying high weights to initial, noise-dominated latents can destabilize training \cite{karras2022edm}. BiasMap acts as a diffusion-aware guidance mechanism: its influence is strategically scaled across diffusion timesteps to align supervision with the progressive denoising process. This ensures that strong guidance is applied only when the signal is reliable, thereby mitigating data bias without sacrificing training stability. Together, LFM and BiasMap enable VistaDepth to achieve a more balanced focus, leading to superior reconstruction in the far range. In this paper, we mitigate the challenge of uneven depth distribution, particularly for far-range perception. Our method, VistaDepth, sets a new state-of-the-art for diffusion-based MDE across five diverse datasets, achieving a 5.2\% AbsRel error on indoor NYUv2 and 91.8\% $\delta_1$ accuracy on outdoor KITTI. Crucially, it delivers a +3.0\% $\delta_1$ gain in data-sparse far-range regions. In summary, our key contributions are:
\begin{itemize}
  \item We introduce LFM, a novel frequency-domain enhancement module that refines latent features through adaptive spectral responses, significantly improving detail preservation in depth reconstruction.
  \item We introduce BiasMap, a diffusion-aware strategy that dynamically counteracts the long-tail data bias by aligning supervision with the iterative denoising process.
  \item VistaDepth achieves SOTA performance for diffusion-based MDE methods, with significant gains in far‑range accuracy.
\end{itemize}
\section{Related Work}
\subsection{Zero-Shot Monocular Depth Estimation}

Zero-shot MDE targets robust performance across novel domains. Approaches to improve generalization have largely followed two paradigms. The first is data-driven, leveraging large-scale, diverse photo collections. This line of work progressed through influential models like MiDaS~\cite{midas} and culminated in methods like DepthAnything~\cite{depthanything, depthanythingV2} and \textcolor{blue}{MoGe~\cite{moge2024}}, which achieved state-of-the-art generalization by training on massive datasets (e.g., 62M images). However, the prohibitive computational cost and reliance on vast, curated data limit the scalability of this approach. 

To address these issues, a more recent model-driven paradigm has emerged, leveraging the powerful generative priors of pre-trained diffusion models~\cite{rombach2022high}. Marigold~\cite{marigold} pioneered this by fine-tuning a Stable Diffusion backbone on only 74K synthetic samples, demonstrating remarkable data efficiency. This model-driven approach has since been advanced by methods like DepthFM~\cite{depthfm} and GeoWizard~\cite{geowizard}, while related works like LRM~\cite{lrm2024} have explored large-scale transformers for direct single-image 3D reconstruction. However, despite their success, these generative methods consistently struggle with the accurate prediction of far-range structures. 

\subsection{Long-Tail Distribution in Depth Estimation}

This persistent failure stems from the long-tail distribution of depth values~\cite{nyuv2, kitti}, a well-documented challenge where abundant near-range samples dominate the training process at the expense of sparse far-range data. Prior work in the discriminative paradigm attempted to mitigate this via two main strategies: re-engineering the loss function to up-weight far-range regions~\cite{jiao2018look}, or using specialized architectures with experts dedicated to distinct depth intervals~\cite{dme}. However, these static strategies are fundamentally incompatible with the dynamic, iterative nature of the diffusion denoising process. Their fixed weights or architectural splits are ill-suited for a process where the signal-to-noise ratio changes dramatically over time. This necessitates a new class of solutions capable of dynamically rebalancing supervision in a manner that is compatible with the generative process, motivating the core contributions of our work.
\section{Methodology}
\subsection{Problem Formulation}
\label{sec:problem_formulation}
Given an input image $\mathbf{x} \in \mathbb{R}^{3\times H\times W}$, the objective is to estimate a corresponding depth map $\mathbf{d} \in \mathbb{R}^{1\times H\times W}$. Depth scales vary by orders of magnitude across datasets, ranging from millimetres in indoor scenes to kilometres in outdoor environments; this dramatic variation can cause models trained on one scale to fail catastrophically when applied to another. To address this challenge, we follow Marigold \cite{marigold} and adopt an affine-invariant normalisation scheme based on percentile statistics. Specifically, given the 2\% and 98\% points \((d_{2}, d_{98})\) of the depth distribution, we apply
\begin{equation}
\tilde{\mathbf{d}}
= 2\!\left(\frac{\mathbf{d}-d_{2}}{d_{98}-d_{2}}-\frac{1}{2}\right),
\end{equation}
which maps the central 96\% of depth values to the interval \([-1,1]\). This percentile-based approach offers three advantages: (i) robustness to outliers that would otherwise dominate traditional min–max normalisation, (ii) preservation of the scene-specific depth distribution, and (iii) compatibility with the input range expected by the frozen, Stable Diffusion pre-trained variational auto-encoder (VAE) \cite{rombach2022high}, preventing numerical instabilities during the diffusion process.

\begin{figure}[h]
  \centering
  \includegraphics[width=\columnwidth, height=0.17\textheight]{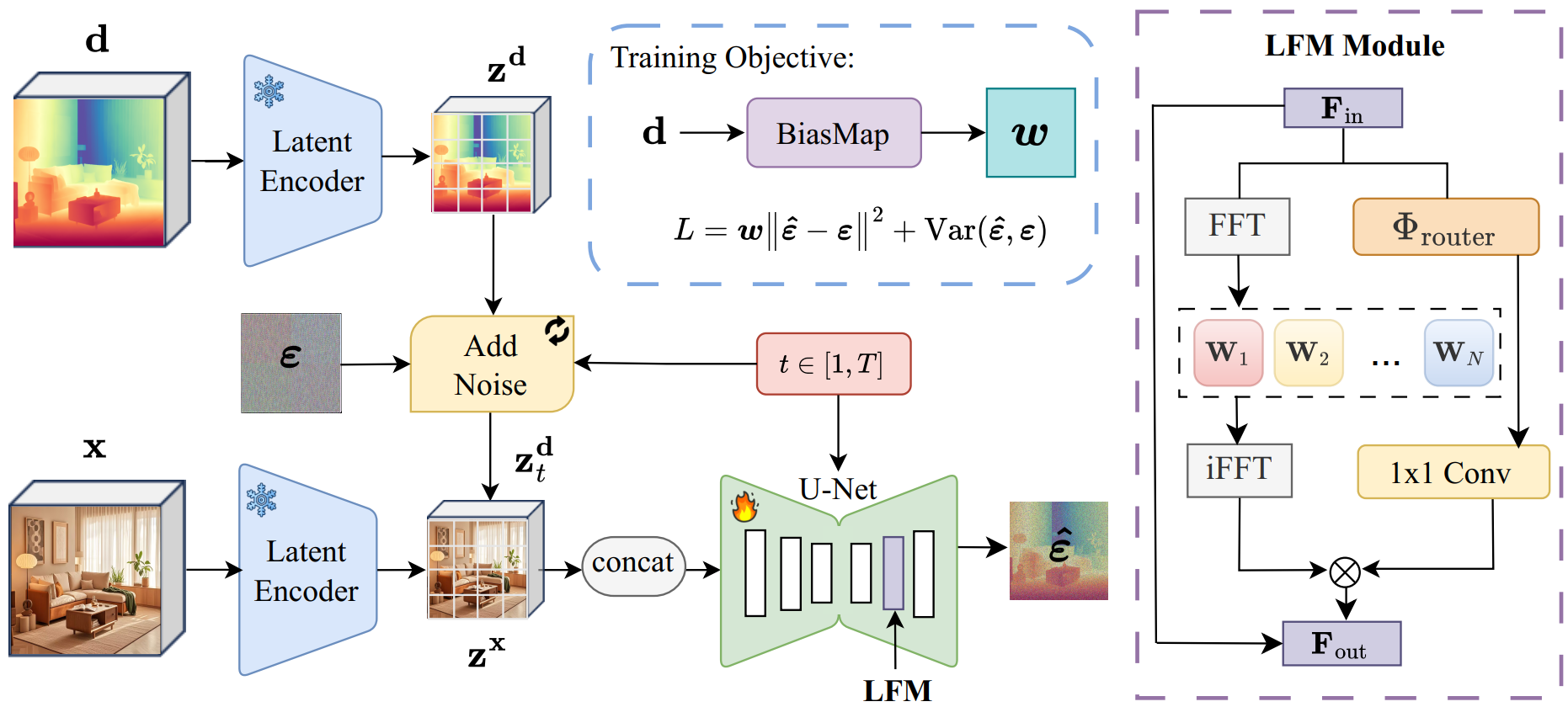}
  \vspace{-0.5cm}
           \caption{\textbf{VistaDepth Training Pipeline.} 
        An RGB image $\mathbf{x}$ and its depth map $\mathbf{d}$ are encoded into latents $\mathbf{z}^{x}$ and $\mathbf{z}_{0}^{d}$. 
        Noise $\boldsymbol{\epsilon}_t$ is added to the depth latent for a random timestep $t$. 
        The LFM-enhanced U-Net takes the concatenated pair $(\mathbf{z}^{x}, \mathbf{z}_{t}^{d})$ as input and is trained to predict the added noise. 
        The LFM module, embedded in the decoder, refines features with content-aware frequency modulation (see Sec.~\ref{sec:Latent-Frequency-Modulation}).}
  \label{Fig:Training_Strategies}
\end{figure}

We frame MDE as learning the conditional distribution \(p(\mathbf{d}\,|\,\mathbf{x})\) with a latent-diffusion probabilistic model \cite{ho2020denoising}. Specifically, we leverage a VAE encoder \(\mathcal{E}(\cdot)\) to transform both the RGB image and its corresponding depth map into a compact latent space:
\begin{equation}
\mathbf{z}^{x} = \mathcal{E}(\mathbf{x}), \quad
\mathbf{z}_0^{d} = \mathcal{E}(\tilde{\mathbf{d}}),
\end{equation}
where \(\mathbf{z}^{x}, \mathbf{z}_0^{d} \in \mathbb{R}^{c\times h\times w}\) denote the latent representations of the RGB image and the normalised depth map, respectively.

The forward diffusion process progressively adds noise to the clean latent \(\mathbf{z}_0^{d}\) over \(T\) discrete time steps to produce a noisy latent \(\mathbf{z}_t^{d}\), following:
\begin{equation}
\mathbf{z}_t^{d}
  = \sqrt{\bar{\alpha}_t}\,\mathbf{z}_0^{d}
  + \sqrt{1-\bar{\alpha}_t}\,\boldsymbol{\epsilon}_t,
\qquad
\boldsymbol{\epsilon}_t \sim \mathcal{N}(\mathbf{0},\mathbf{I}),
\end{equation}
where \(\bar{\alpha}_t\) is the cumulative product of the variance-schedule parameters. The reverse process is learned by a U-Net, enhanced with our proposed LFM module, which is trained to predict the noise \(\hat{\boldsymbol{\epsilon}}_{t}=\boldsymbol{\epsilon}_{\theta}(\mathbf{z}_t^{d},\mathbf{z}^{x},t)\). During inference, this prediction is used to iteratively recover the clean latent \(\hat{\mathbf{z}}_0^{d}\) from pure noise \(\mathbf{z}_T^{d}\) via the DDIM sampler \cite{song2020denoising}.

\subsection{Overview of VistaDepth Architecture}
Our framework, VistaDepth, enhances a standard latent diffusion model with two complementary modules operating at the feature and supervisory levels, respectively (Fig.~\ref{Fig:Training_Strategies}). It is built upon a frozen VAE and a trainable U-Net backbone from the pre-trained Stable Diffusion \cite{rombach2022high}. At the feature level, we introduce the LFM module into the U-Net decoder. Its purpose is to counteract the backbone's spectral bias and improve the preservation of fine geometric details, particularly for distant structures. At the supervisory level, we employ the BiasMap mechanism. This module dynamically reweights the diffusion loss during training to rebalance focus towards under-represented far-range and structurally complex regions. This mechanism is disabled during inference. This dual-pronged approach of refining features internally while guiding supervision externally enables a more robust and balanced learning process for MDE.

\begin{figure}[h]
  \centering
  \includegraphics[width=\columnwidth, height=0.14\textheight]{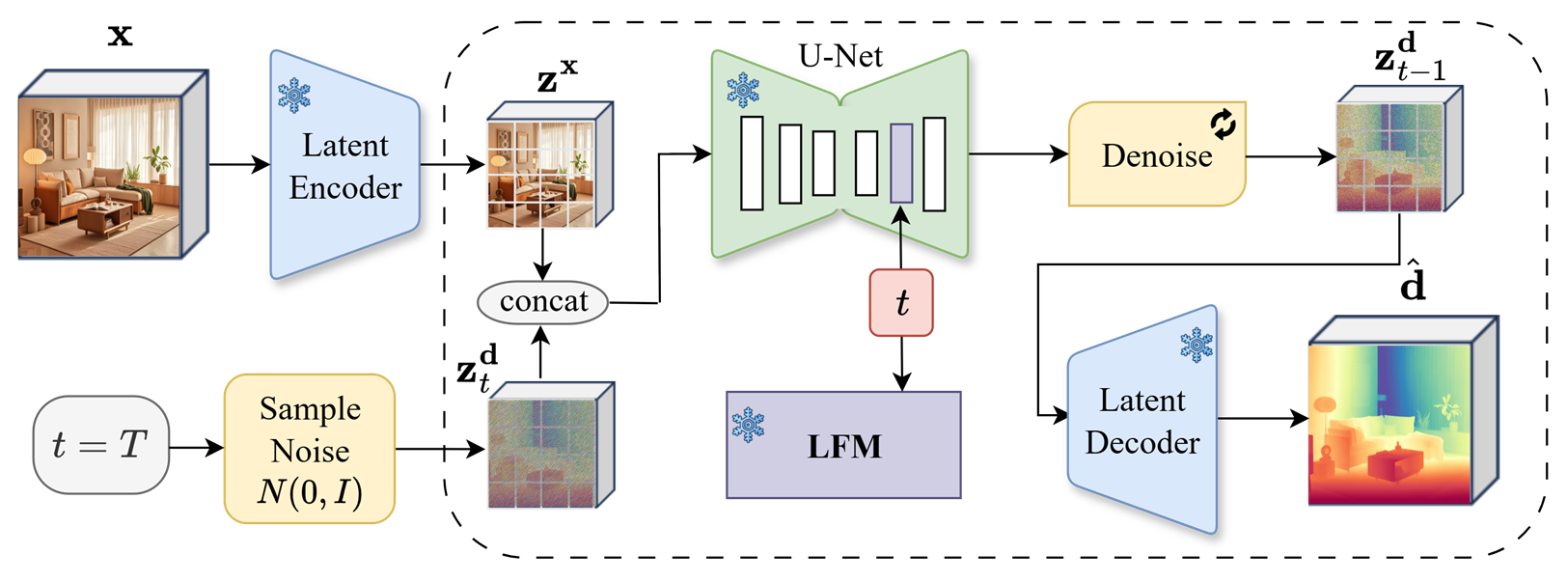}
  \vspace{-0.6cm}
   \caption{
      \textbf{VistaDepth Inference Pipeline.} 
      An input RGB image $\mathbf{x}$ is encoded into a conditioning latent $\mathbf{z}^x$. 
      Separately, a tensor of pure Gaussian noise $\mathbf{z}_T^{d}$ is iteratively denoised by the LFM-enhanced U-Net. 
      At each step, the network predicts the noise from the current latent $\mathbf{z}_t^{d}$, conditioned on $\mathbf{z}^x$. 
      The final clean latent $\hat{\mathbf{z}}_0^{d}$ is passed through the VAE decoder to reconstruct the high-resolution depth map.
  }
  \label{fig:inference_pipeline}
\end{figure}

\subsection{Training and Inference Procedures}
\label{sec:training_and_inference}

The training procedure, illustrated in Figure~\ref{Fig:Training_Strategies}, is designed to teach our LFM-enhanced U-Net to reverse the diffusion process. For each training sample, we first encode the RGB image and its ground-truth depth map into latent representations. A noisy latent is then created by corrupting the clean depth latent with Gaussian noise for a random timestep $t$. The U-Net is trained to predict this noise, conditioned on the image latent via channel-wise concatenation. To adapt the U-Net's input layer for this expanded input, we duplicate and scale its pre-trained weights. The network is optimized using the loss function detailed in Sec.~\ref{sec:loss}, dynamically weighted by our BiasMap mechanism.

The inference pipeline, illustrated in Figure~\ref{fig:inference_pipeline}, then uses this trained network to generate a depth map from a single RGB image. Starting from pure Gaussian noise, our trained U-Net iteratively denoises the latent over a predefined number of timesteps, conditioned on the image latent. The BiasMap module is disabled during this stage. The final clean latent is then transformed by the VAE decoder into the high-resolution depth map.

\subsection{Latent Frequency Modulation}
\label{sec:Latent-Frequency-Modulation}
The LFM module is a feature refinement block inserted into the U-Net decoder. It enhances features through an internal frequency-domain sub-pathway that performs spatially-adaptive spectral modulation, addressing a known challenge for standard convolutions in high-frequency tasks \cite{ye2024diffusionedge}. The LFM module first transforms the input features $\mathbf{F}_{\text{in}} \in \mathbb{R}^{c \times h \times w}$ into the frequency domain using a 2D Fast Fourier Transform (FFT), and decomposes the resulting complex tensor into its magnitude $\mathbf{A}$ and phase $\mathbf{P}$. Our modulation exclusively targets the magnitude while deliberately preserving the phase.

The core of the LFM's mechanism involves two synergistic components: a bank of $N$ learnable global spectral masks, $\mathbf{W}_{\text{bank}}$, and a content-aware routing network, $\Phi_{\text{router}}$. Each mask in $\mathbf{W}_{\text{bank}}$ learns a distinct global frequency response, forming a basis set of spectral filters. The routing network processes the spatial features $\mathbf{F}_{\text{in}}$ to generate a spatially-variant mixing map, $\mathbf{S}_{\text{select}}$, that locally combines these basis filters. Generating a high-fidelity mixing map requires multi-scale spatial reasoning. Therefore, its architecture comprises two sequential $3 \times 3$ convolutional layers for local pattern extraction, a multi-head self-attention layer~\cite{vaswani2017attention} for global context aggregation, and a final $1 \times 1$ convolutional layer for projection, followed by a per-pixel softmax. The $N$ spectral masks are applied to the magnitude spectrum $\mathbf{A}$. Each modulated magnitude is then recombined with the original phase $\mathbf{P}$ and transformed back to the spatial domain via an Inverse FFT, yielding a set of $N$ candidate features:
\begin{equation}
  \mathbf{F}_{\text{filtered}, i} = \mathcal{F}^{-1}\!\bigl(( \mathbf{W}_{\text{bank}, i} \odot \mathbf{A} ) \cdot e^{i\mathbf{P}}\bigr).
\end{equation}
These candidates are then synthesized in the spatial domain using the mixing weights from $\mathbf{S}_{\text{select}}$:
\begin{equation}
  \mathbf{F}_{\text{mixed}} = \sum_{i=1}^{N} \mathbf{S}_{\text{select}, i} \odot \mathbf{F}_{\text{filtered}, i},
\end{equation}
where $\mathbf{S}_{\text{select}, i}$ is broadcast across channels. This synthesis operation effectively constructs a spatially-variant filter by linearly combining the basis filters. The final spatially-refined feature is integrated back into the main decoder path via a residual connection:
\begin{equation}
  \mathbf{F}_{\text{out}} = \mathbf{F}_{\text{in}} + \mathbf{F}_{\text{mixed}}.
\end{equation}

\subsection{BiasMap Reweighting Mechanism}
\label{sec:biasmap}
To steer the model's focus towards perceptually critical far-range signals over uninformative low-frequency areas, BiasMap decomposes supervision difficulty into two complementary components: distance and structural complexity.

\textbf{Spatial Weight Decomposition.}
Following the affine-normalised depth map $\tilde{\mathbf{d}}$ from Sec.~\ref{sec:problem_formulation}, the distance-based weight is defined as $\boldsymbol{w}_{\text{dist}} = (\tilde{\mathbf{d}} + 1) / 2$, which linearly upweights far-range regions. The structural complexity, $\boldsymbol{w}_{\text{struct}}$, is the normalised gradient magnitude of $\tilde{\mathbf{d}}$, emphasizing regions with high depth discontinuities. To apply these weights in the latent space, they are first downsampled by 8 to match the latent resolution. We employ an asymmetric pooling strategy: average pooling for the smooth $\boldsymbol{w}_{\text{dist}}$ to preserve its regional statistics and max pooling for the sparse $\boldsymbol{w}_{\text{struct}}$ to retain peak edge signals. The downsampled maps, $\bar{\boldsymbol{w}}_{\text{dist}}$ and $\bar{\boldsymbol{w}}_{\text{struct}}$, are then fused using a learnable soft gating unit:
\begin{equation}
\boldsymbol{g}
  = \sigma\!\bigl(\bar{\boldsymbol{w}}_{\text{dist}}\odot\bar{\boldsymbol{w}}_{\text{struct}}-\tau\bigr),
\end{equation}
where $\tau$ is a learnable bias. This formulation is superior because a simple element-wise product can suppress important signals if one weight is near zero. The learnable bias $\tau$ allows the model to dynamically adjust the gate's activation threshold. This enables it to learn a more nuanced fusion strategy, such as emphasizing regions where both signals are strong, or where one is moderate while the other is critical. The final spatial weight map $\boldsymbol{w}$ is produced by normalising $\boldsymbol{g}$ batch-wise:
\begin{equation}
\boldsymbol{w}
  = \frac{\boldsymbol{g}}{\langle\boldsymbol{g}\rangle+\kappa},
\end{equation}
where $\langle\cdot\rangle$ is the batch-wise mean and $\kappa=10^{-6}$.

\begin{figure*}[t]
    \centering
    \includegraphics[width=0.95\textwidth, keepaspectratio]{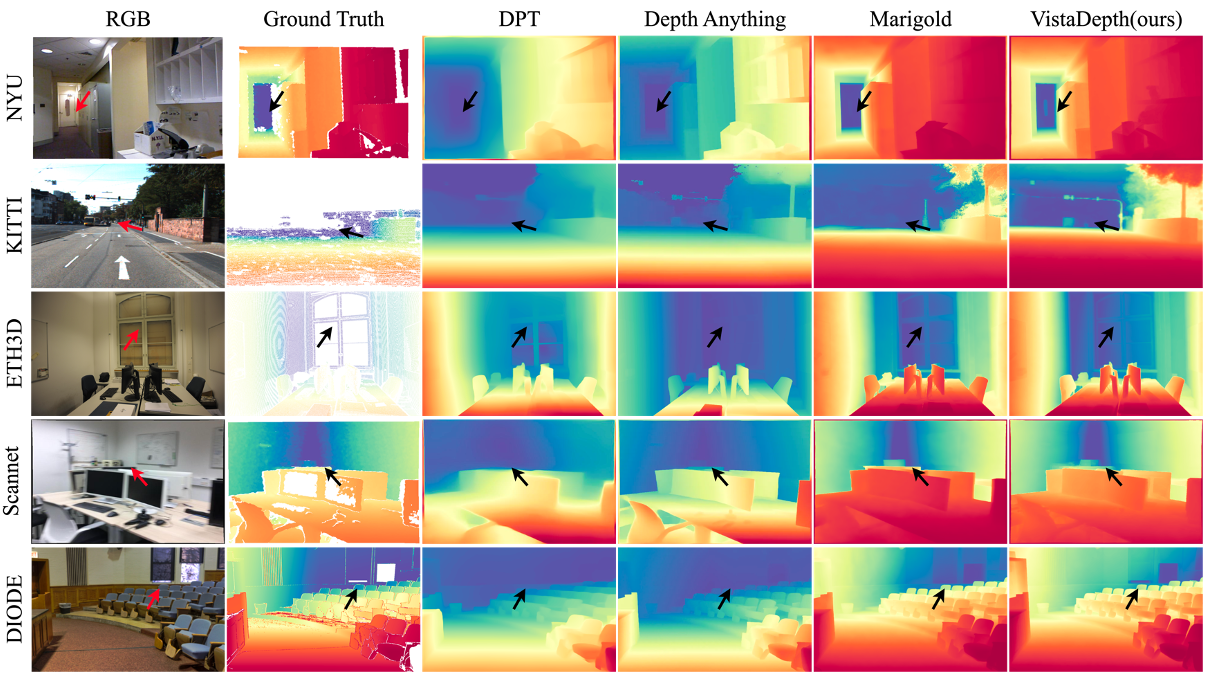}
    \vspace{-0.2cm}
    \caption{
    Qualitative comparison of MDE across different datasets. VistaDepth demonstrates superior performance in capturing fine details at a distance (e.g., edges of doorways in NYU, distant cars in KITTI, windows in ETH3D and DIODE) and maintaining scene consistency (e.g., complex interior scene in ScanNet).
    }
    \label{fig:qualitative_comparisons}
\end{figure*}

\textbf{SNR-Aware Temporal Modulation.}
A static application of $\boldsymbol{w}$ is ill-suited for the dynamic denoising process, as applying strong weights to early, noise-dominated latents can destabilize training. We therefore introduce an SNR-aware temporal modulation schedule that acts as a form of curriculum learning. The instantaneous SNR at timestep $t$, $\operatorname{SNR}_t = \bar{\alpha}_t / (1-\bar{\alpha}_t)$, drives a monotonic ramp factor $\eta_t = (\operatorname{SNR}_t/\operatorname{SNR}_{\max})^{\gamma}$, where we set $\gamma=5$ based on a sensitivity analysis (Fig.~\ref{fig:time_period}). The final time-aware weights are a convex blend between uniform weighting and full BiasMap guidance:
\begin{equation}
\boldsymbol{w}_{\text{final},t}
=(1-\eta_t)\,\mathbf{1}+\eta_t\,\boldsymbol{w}.
\label{eq:w_final}
\end{equation}
This ensures strong guidance is applied only when the signal is reliable, mitigating data bias without sacrificing training stability.

\subsection{Loss Function}
\label{sec:loss}
The primary training objective is a re-weighted squared error in the latent space, where the error between the predicted noise \(\hat{\boldsymbol{\epsilon}}_t\) and ground-truth noise \(\boldsymbol{\epsilon}_t\) is weighted by our time-aware map \(\boldsymbol{w}_{\text{final},t}\):
\begin{equation}
  \mathcal{L}_{\text{latent}}
  = \frac{1}{M}\sum_{i=1}^{M}
    \boldsymbol{w}_{\text{final},t,i}\,
    \bigl(\hat{\boldsymbol{\epsilon}}_{t,i}-\boldsymbol{\epsilon}_{t,i}\bigr)^{2},
\end{equation}
where \(M\) is the number of latent elements. However, relying solely on $\mathcal{L}_{\text{latent}}$ risks a common pitfall of re-weighting schemes \cite{lin2017focal}: focusing only on high-weight areas. This is particularly severe for diffusion models, where unconstrained errors in low-weight regions can amplify into visual artifacts during iterative denoising. To mitigate this, we introduce a variance regulariser that directly penalises the dispersion of the prediction error, preventing the model from achieving a low mean error by simply offsetting large positive and negative errors:
\begin{equation}
  \mathcal{L}_{\text{var}}
  = \operatorname{Var}_{i}\!\bigl(\hat{\boldsymbol{\epsilon}}_{t,i}-\boldsymbol{\epsilon}_{t,i}\bigr).
  \label{eq_loss_var}
\end{equation}
The overall objective is a weighted combination of these two terms:
\begin{equation}
  \mathcal{L}_{\text{total}}
  = \mathcal{L}_{\text{latent}} + \lambda\,\mathcal{L}_{\text{var}}.
  \label{eq_loss_total}
\end{equation}
We set the hyper-parameter $\lambda=1$ based on a grid search over \(\{0.01, 0.1, 1, 10\}\) on a held-out validation set, as it provides the best trade-off between far-range detail accuracy and overall prediction consistency.

\section{Experiments}
\subsection{Implementation Details}
We initialize our backbone network using the pre-trained weights from Stable Diffusion \cite{rombach2022high} and fine-tune only the U-Net. The latent diffusion backbone uses a pre-trained VAE with a downsampling factor of \( r = 8 \) and \( c = 4 \) channels. Meanwhile, the LFM module operates in this latent space. During training, we employ the DDPM noise scheduler with 1000 diffusion steps, while at inference, the DDIM noise scheduler \cite{song2020denoising} is used with 50 steps. The final predictions are obtained by averaging the results across 8 inference runs, each with a different noise seed. For a fair and direct comparison, we apply this same multi-run averaging protocol to all diffusion-based methods in our benchmark tables, including the official Marigold implementation. The model is trained for 14,000 steps using the AdamW optimizer, with a learning rate of \( 3 \times 10^{-5} \) and a batch size of 8 by default. Our model takes one day to train using a single NVIDIA A100 GPU.

\subsection{Benchmarking and Comparative Evaluation}
\label{sec:datasets_and_metrics}

\textbf{Training datasets.}
We follow Marigold \cite{marigold} and use 74K samples from two synthetic datasets, Hypersim \cite{hypersim} and Virtual KITTI \cite{vkitti}. Hypersim consists of 54K samples of indoor scenes with RGB images and depth maps resized to $480 \times 640$. Virtual KITTI provides 20K samples of outdoor driving scenes across four scenarios with varied conditions.

\textbf{Evaluation Datasets and Protocol.}
We evaluate our model on five real-world datasets that were not used in training: NYUv2 \cite{nyuv2}, an indoor dataset with 654 images; ScanNet \cite{scannet}, where we randomly sample 800 images from 312 validation scenes; KITTI \cite{kitti}, an outdoor driving dataset with sparse LiDAR depth, evaluated on the Eigen split with 652 images; ETH3D \cite{eth3d}, evaluated on all 454 samples with ground truth; and DIODE \cite{diode}, which covers both indoor and outdoor scenes, tested on 325 indoor and 446 outdoor images. The model’s performance is assessed using two metrics: Absolute Mean Relative Error (AbsRel), defined as 
\( \frac{1}{N} \sum_{i=1}^{N} \left| \frac{\hat{d}_i - d_i}{d_i} \right| \)
and \( \delta_1 \) accuracy, which measures the percentage of pixels satisfying \( \max \left( \frac{\hat{d}_i}{d_i}, \frac{d_i}{\hat{d}_i} \right) < 1.25 \).

\begin{table*}[t]
  \centering
  \caption{Zero-shot depth estimation on five public benchmarks. Metrics are Absolute Relative Error (AbsRel $\downarrow$) and threshold accuracy $\delta_1$ $\uparrow$. The table contrasts discriminative, diffusion-based generative, and our proposed VistaDepth models. \textbf{Bold} and \underline{underline} denote the best and second-best scores within each model family, respectively (lower AbsRel / higher $\delta_1$ are better).}
  \label{tab:comparison_sota}
  \footnotesize
  \resizebox{\textwidth}{!}{%
    \begin{tabular}{c l c c cc cc cc cc cc}
      \toprule
      & \multirow{2}{*}{\textbf{Method}} &
      \multirow{2}{*}{\textbf{Train Data}} &
      \multirow{2}{*}{\textbf{Prior}} &
      \multicolumn{2}{c}{NYUv2} &
      \multicolumn{2}{c}{KITTI} &
      \multicolumn{2}{c}{ETH3D} &
      \multicolumn{2}{c}{ScanNet} &
      \multicolumn{2}{c}{DIODE} \\
      \cmidrule(lr){5-6}\cmidrule(lr){7-8}\cmidrule(lr){9-10}\cmidrule(lr){11-12}\cmidrule(lr){13-14}
      & & & & AbsRel $\downarrow$ & $\delta_1$ $\uparrow$ & AbsRel $\downarrow$ & $\delta_1$ $\uparrow$ &
        AbsRel $\downarrow$ & $\delta_1$ $\uparrow$ & AbsRel $\downarrow$ & $\delta_1$ $\uparrow$ &
        AbsRel $\downarrow$ & $\delta_1$ $\uparrow$ \\[0.1em]
      \midrule
      \multirow{10}{*}{\rotatebox[origin=c]{90}{\textcolor{gray}{\textit{Discriminative (data-driven)}}}}
      & \textcolor{gray}{MiDaS \cite{midas}}                      
      & \textcolor{gray}{2\,M}    
      & \textcolor{gray}{ImageNet} 
      & \textcolor{gray}{9.5}  
      & \textcolor{gray}{91.5} 
      & \textcolor{gray}{18.3} 
      & \textcolor{gray}{71.1} 
      & \textcolor{gray}{19.0} 
      & \textcolor{gray}{88.4} 
      & \textcolor{gray}{9.9}  
      & \textcolor{gray}{90.7} 
      & \textcolor{gray}{26.6} 
      & \textcolor{gray}{71.3} \\
      & \textcolor{gray}{LeReS \cite{yin2021learning}}            
      & \textcolor{gray}{354\,K}  
      & \textcolor{gray}{ImageNet} 
      & \textcolor{gray}{9.0}  
      & \textcolor{gray}{91.6} 
      & \textcolor{gray}{14.9} 
      & \textcolor{gray}{78.4} 
      & \textcolor{gray}{17.1} 
      & \textcolor{gray}{77.7} 
      & \textcolor{gray}{9.1}  
      & \textcolor{gray}{91.7} 
      & \textcolor{gray}{27.1} 
      & \textcolor{gray}{76.6} \\
      & \textcolor{gray}{Omnidata \cite{eftekhar2021omnidata}}    
      & \textcolor{gray}{12.2\,M} 
      & \textcolor{gray}{ImageNet} 
      & \textcolor{gray}{7.4}  
      & \textcolor{gray}{94.5} 
      & \textcolor{gray}{14.9} 
      & \textcolor{gray}{83.5} 
      & \textcolor{gray}{16.6} 
      & \textcolor{gray}{77.8} 
      & \textcolor{gray}{7.5}  
      & \textcolor{gray}{93.6} 
      & \textcolor{gray}{33.9} 
      & \textcolor{gray}{74.2} \\
      & \textcolor{gray}{DPT \cite{DPT}}                          
      & \textcolor{gray}{1.4\,M}  
      & \textcolor{gray}{ImageNet} 
      & \textcolor{gray}{9.1}  
      & \textcolor{gray}{91.9} 
      & \textcolor{gray}{11.1} 
      & \textcolor{gray}{88.1} 
      & \textcolor{gray}{11.5} 
      & \textcolor{gray}{92.9} 
      & \textcolor{gray}{8.4}  
      & \textcolor{gray}{93.2} 
      & \textcolor{gray}{26.9} 
      & \textcolor{gray}{73.0} \\
      & \textcolor{gray}{Lotus-D \cite{lotus}}                    
      & \textcolor{gray}{59\,K}   
      & \textcolor{gray}{SD v2.0}  
      & \textcolor{gray}{5.3}  
      & \textcolor{gray}{96.7} 
      & \textcolor{gray}{9.3}  
      & \textcolor{gray}{92.8} 
      & \textcolor{gray}{6.8}  
      & \textcolor{gray}{95.3} 
      & \textcolor{gray}{6.0}  
      & \textcolor{gray}{96.1} 
      & \textcolor{gray}{22.8} 
      & \textcolor{gray}{73.8} \\
      & \textcolor{gray}{Depth Anything \cite{depthanything}}     
      & \textcolor{gray}{63.5\,M} 
      & \textcolor{gray}{DINOv2}   
      & \textcolor{gray}{4.3}  
      & \textcolor{gray}{\underline{98.1}} 
      & \textcolor{gray}{8.0}  
      & \textcolor{gray}{94.6} 
      & \textcolor{gray}{12.7} 
      & \textcolor{gray}{88.2} 
      & \textcolor{gray}{\underline{4.3}}  
      & \textcolor{gray}{\textbf{98.1}} 
      & \textcolor{gray}{27.7} 
      & \textcolor{gray}{75.9} \\
      & \textcolor{gray}{Depth Anything v2 \cite{depthanythingV2}}
      & \textcolor{gray}{63.5\,M} 
      & \textcolor{gray}{DINOv2}   
      & \textcolor{gray}{4.4}  
      & \textcolor{gray}{97.9} 
      & \textcolor{gray}{7.5}  
      & \textcolor{gray}{94.8} 
      & \textcolor{gray}{13.2} 
      & \textcolor{gray}{86.2} 
      & \textcolor{gray}{\textbf{4.2}}  
      & \textcolor{gray}{\underline{97.8}} 
      & \textcolor{gray}{26.5} 
      & \textcolor{gray}{73.4} \\
      & \textcolor{gray}{Metric3D \cite{metric3d}}                
      & \textcolor{gray}{8\,M}    
      & \textcolor{gray}{ImageNet} 
      & \textcolor{gray}{5.8}  
      & \textcolor{gray}{96.3} 
      & \textcolor{gray}{5.3}  
      & \textcolor{gray}{96.5} 
      & \textcolor{gray}{6.4}  
      & \textcolor{gray}{96.5} 
      & \textcolor{gray}{7.4}  
      & \textcolor{gray}{94.2} 
      & \textcolor{gray}{21.1} 
      & \textcolor{gray}{82.5} \\
      & \textcolor{gray}{Metric3D v2 \cite{metric3dv2}}           
      & \textcolor{gray}{16\,M}   
      & \textcolor{gray}{DINOv2}   
      & \textcolor{gray}{\underline{4.3}}  
      & \textcolor{gray}{\underline{98.1}} 
      & \textcolor{gray}{\underline{4.4}}  
      & \textcolor{gray}{\textbf{98.2}} 
      & \textcolor{gray}{\underline{4.2}}  
      & \textcolor{gray}{\underline{98.3}} 
      & \textcolor{gray}{--}   
      & \textcolor{gray}{--}   
      & \textcolor{gray}{\underline{13.6}} 
      & \textcolor{gray}{\underline{89.5}} \\
      & \textcolor{gray}{MoGe \cite{moge2024}}                    
      & \textcolor{gray}{9\,M}   
      & \textcolor{gray}{DINOv2}   
      & \textcolor{gray}{\textbf{2.9}} 
      & \textcolor{gray}{\textbf{98.6}} 
      & \textcolor{gray}{\textbf{3.9}} 
      & \textcolor{gray}{\underline{98.0}} 
      & \textcolor{gray}{\textbf{2.6}} 
      & \textcolor{gray}{\textbf{99.2}} 
      & \textcolor{gray}{--}   
      & \textcolor{gray}{--}   
      & \textcolor{gray}{\textbf{3.1}} 
      & \textcolor{gray}{\textbf{97.5}} \\
      \midrule
      \multirow{7}{*}{\rotatebox[origin=c]{90}{\textit{Generative}}}
      & Marigold \cite{marigold}                & 74\,K   & SD v2.0  & 5.5  & 96.4 & 10.5 & 90.2 & 6.5  & 96.0 & 6.4  & 95.1 & 30.8 & 77.3 \\
      & Marigold LCM \cite{marigoldlcm}         & 74\,K   & SD v2.0  & 5.8  & 96.1 & 10.1 & 90.9 & 6.8  & 95.6 & 6.6  & 95.0 & 30.5 & 77.2 \\
      & GeoWizard \cite{geowizard}              & 280\,K  & SD v2.0  & 5.6  & 96.3 & 10.3 & 89.2 & 6.4  & 95.8 & 6.1  & 95.3 & 30.6 & 77.6 \\
      & DepthFM \cite{depthfm}                  & 63\,K   & SD v2.0  & 6.5  & 95.6 & 10.1 & 90.9 & 9.4  & 90.9 & 8.5  & 92.3 & \underline{25.1} & \textbf{78.3} \\
      & GenPercept \cite{GenPercept}            & 74\,K   & SD v2.0  & 5.6  & 96.0 & 9.9  & 90.4 & 6.9  & 95.6 & 6.2  & 95.9 & 35.7 & 75.6 \\
      & Lotus-G \cite{lotus}                    & 59\,K   & SD v2.0  & \underline{5.4} & \underline{96.6} & 11.3 & 87.7 & \underline{6.2} & \underline{96.1} & 6.0  & 96.0 & \textbf{22.9} & 72.9 \\
      & E2E-FT \cite{e2e}                       & 74\,K   & SD v2.0  & \underline{5.4} & 96.5 & \underline{9.8} & \underline{91.2} & 6.4  & 95.9 & \textbf{5.8} & \underline{96.2} & 30.3 & 77.6 \\
      \cmidrule(lr){2-14}
      & \textbf{VistaDepth (Ours)}              & 74\,K   & SD v2.0  & \textbf{5.2} & \textbf{97.4} & \textbf{9.7} & \textbf{91.8} & \textbf{6.1} & \textbf{96.4} & \underline{5.9} & \textbf{96.3} & 28.6 & \underline{78.1} \\
      \bottomrule
    \end{tabular}%
  }
\end{table*}

\subsection{Comparisons with Other Models}
\label{sec:Comparisons with Other Models}
We compare VistaDepth against state-of-the-art methods, categorized into diffusion-based generative models and large-scale discriminative networks, with full results presented in Table~\ref{tab:comparison_sota}. The results underscore the strength of diffusion-based approaches, which achieve remarkable generalization from significantly smaller training sets. Within this paradigm, VistaDepth sets a new state-of-the-art on a majority of the benchmarks, including NYUv2, KITTI, and ETH3D. Notably, the data efficiency of our approach is particularly stark when contrasted with the discriminative paradigm. For instance, VistaDepth surpasses the performance of Depth Anything v2~\cite{depthanythingV2} on the challenging ETH3D benchmark, while being fine-tuned on approximately 850 times less data (74K synthetic pairs vs. 63.5M labeled images). To ensure a fair comparison, the ScanNet benchmark was evaluated on a publicly available data split, as some prior works utilized inaccessible splits.

\subsection{Ablation Studies}
\label{sec:Ablation Studies} 
\textbf{Depth Range Evaluation.}
To validate VistaDepth under long-tailed distributions, we conducted ablation studies by segmenting the test datasets into four depth ranges to evaluate performance across varying depths. As shown in Table \ref{tab:depth_ranges_no_avg}, when compared to other leading diffusion-based methods, our method not only maintains excellent performance in the near-range but also significantly improves accuracy in the far-range.
\begin{table}[h!]
  \centering
  \small
\caption{Performance breakdown by depth range ($\delta_1$$\uparrow$). This experiment validates VistaDepth's significant advantage in the challenging, data-sparse far-range regions.}
  \label{tab:depth_ranges_no_avg}
  \begin{tabular}{l cccc}
    \toprule
    \multirow{2.5}{*}{Method} & \multicolumn{4}{c}{Depth Range ($\delta_1$ Accuracy $\uparrow$)} \\
    \cmidrule(lr){2-5}
    & 0--20\% & 20--40\% & 40--60\% & Beyond 60\% \\
    \midrule
    Marigold \cite{marigold}   & \underline{91.2} & 96.1 & 90.0 & 75.3 \\
    GeoWizard \cite{geowizard} & 89.6 & \underline{97.5} & 88.9 & 76.1 \\
    GenPercept \cite{GenPercept} & \underline{91.2} & 95.5 & \underline{92.0} & 74.5 \\
    Lotus-G \cite{lotus}       & 89.0 & 97.1 & 90.0 & 76.2 \\
    E2E-FT \cite{e2e} & 91.1 & \underline{97.5} & 88.4 & \underline{76.6} \\
    \midrule
    \textbf{VistaDepth (Ours)} & \textbf{91.2} & \textbf{97.9} & \textbf{92.8} & \textbf{79.6} \\
    \textbf{Gain over Best}    & \textcolor{gray}{0.0} & \textcolor{blue!70!black}{+0.4} & \textcolor{blue!70!black}{+0.8} & \textcolor{red!70!black}{\textbf{+3.0}} \\
    \bottomrule
  \end{tabular}
\end{table}

\textbf{Ablation on BiasMap Components.}
To isolate the impact of BiasMap, we validate the additive contribution of each of its primary components with the LFM module disabled. As shown in Table~\ref{tab:ablation_biasmap}, each component yields a clear, incremental performance gain. Enabling the distance weight ($\boldsymbol{w}_{\text{dist}}$) alone improves performance by forcing the model to allocate more capacity to under-represented far-range regions. The structure weight ($\boldsymbol{w}_{\text{struct}}$) provides an even larger boost by focusing supervision on challenging high-frequency details like object boundaries. Finally, adding the variance regulariser ($\mathcal{L}_{\text{var}}$) yields a further improvement, confirming its critical role as a stabilizer that prevents the model from overfitting to highly-weighted regions at the expense of overall prediction quality. This step-by-step validation confirms that all three components are essential and synergistically mitigate supervision bias.
\begin{table}[h!]
  \centering
  \small
  \setlength{\tabcolsep}{8pt}
  \caption{Ablation study validating the components of the BiasMap mechanism. Checkmarks (\ding{51}) and crosses (\ding{55}) denote active and inactive components, respectively.}
  \label{tab:ablation_biasmap}

  \resizebox{\columnwidth}{!}{%
  \begin{tabular}{ccc cc cc}
    \toprule
    \multicolumn{3}{c}{Components} &
    \multicolumn{2}{c}{NYUv2} &
    \multicolumn{2}{c}{KITTI} \\
    \cmidrule(lr){1-3} \cmidrule(lr){4-5} \cmidrule(lr){6-7}
    $\boldsymbol{w}_{\text{dist}}$ & $\boldsymbol{w}_{\text{struct}}$ & $\mathcal{L}_{\text{var}}$ & AbsRel$\downarrow$ & $\delta_1$$\uparrow$ & AbsRel$\downarrow$ & $\delta_1$$\uparrow$ \\ 
    \midrule
    \ding{55} & \ding{55} & \ding{55} & 6.2 & 94.8 & 10.9 & 89.8 \\
    \midrule
    \ding{51} & \ding{55} & \ding{55} & 6.0 & 95.1 & 10.5 & 90.6 \\
    \ding{55} & \ding{51} & \ding{55} & 5.9 & 95.7 & 10.3 & 90.9 \\
    \ding{51} & \ding{51} & \ding{55} & 5.5 & 96.5 & 10.1 & 91.2 \\
    \midrule
    \textbf{\ding{51}} & \textbf{\ding{51}} & \textbf{\ding{51}} & \textbf{5.4} & \textbf{96.7} & \textbf{9.9} & \textbf{91.4} \\
    \bottomrule
  \end{tabular}%
  }
\end{table}

\textbf{Ablation on Asymmetric Pooling.}
We validate our asymmetric pooling strategy for downsampling the spatial weight maps. As shown in Table~\ref{tab:ablation_pooling}, symmetric pooling strategies perform suboptimally due to a fundamental trade-off. Universal average pooling preserves the regional statistics of the smooth distance map ($\boldsymbol{w}_{\text{dist}}$) but dilutes the critical peak signals of the sparse structure map ($\boldsymbol{w}_{\text{struct}}$), losing critical edge information. Conversely, universal max pooling retains these sharp edge signals but fails to capture the representative averages of the distance map, leading to biased supervision. Our asymmetric approach (Average for $\boldsymbol{w}_{\text{dist}}$, Max for $\boldsymbol{w}_{\text{struct}}$) effectively resolves this trade-off, confirming that tailoring the pooling strategy to the signal characteristics of each map is crucial for optimal performance.
\begin{table}
  \centering
\caption{Ablation on pooling strategies for downsampling the BiasMap components.}
  \label{tab:ablation_pooling} 

  \begin{tabular}{cccccc}
    \toprule
    \multirow{2}{*}{$\boldsymbol{w}_{\text{dist}}$} & 
    \multirow{2}{*}{$\boldsymbol{w}_{\text{struct}}$} &
    \multicolumn{2}{c}{NYUv2} &
    \multicolumn{2}{c}{KITTI} \\
    \cmidrule(lr){3-4}\cmidrule(lr){5-6}
    && AbsRel $\downarrow$ & $\delta_1$ $\uparrow$ & AbsRel $\downarrow$ & $\delta_1$ $\uparrow$ \\ 
    \midrule
    Max     & Max     & 5.9 & 95.5 & 10.5 & 90.5 \\
    Max     & Average & 5.9 & 95.6 & 10.4 & 90.6 \\
    Average & Average & 5.8 & 95.7 & 10.3 & 90.8 \\
    
    \textbf{Average} & \textbf{Max} & \textbf{5.4} & \textbf{96.7} & \textbf{9.9} & \textbf{91.4} \\
    \bottomrule
  \end{tabular}
\end{table}

\begin{figure}[t]
\centering
\captionsetup{skip=-0.6pt}
\includegraphics[
width=0.95\linewidth, height=0.14\textheight]{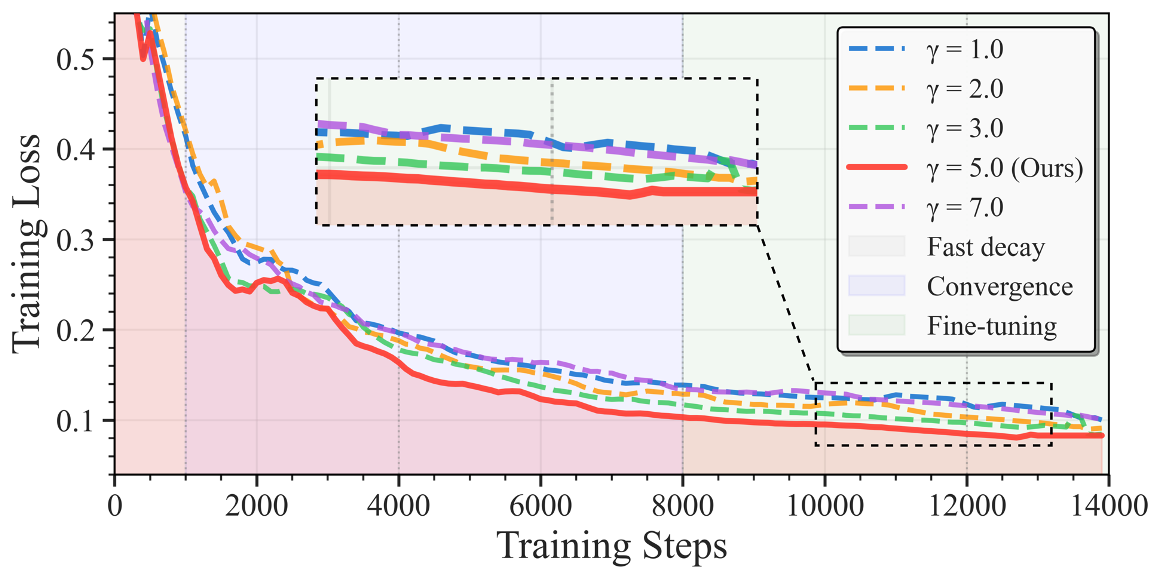}
\caption{Ablation on the temporal modulation $\gamma$. Training loss curves for different values of the ramp factor exponent from Sec.~\ref{sec:biasmap}.}
\label{fig:time_period}
\end{figure}

\textbf{Ablation on SNR-Aware Modulation.}
The hyperparameter $\gamma$ in our SNR-aware temporal modulation, introduced in Sec.~\ref{sec:biasmap}, is critical as it governs the entire training dynamic. Our analysis of the training loss curves in Figure~\ref{fig:time_period} reveals a clear trade-off: a small $\gamma$ (e.g., 1.0) causes training instability by applying weights too aggressively in early, noise-dominated steps, while a large $\gamma$ (e.g., 20.0) dilutes the guidance's impact by delaying it for too long. The analysis identifies $\gamma=5.0$ as the optimal trade-off, leading to stable convergence and strong validation performance.

\textbf{Ablation on LFM Architectural Placement.}
To determine the optimal placement for the LFM module, we insert a single module at various U-Net stages (Table~\ref{tab:lfm_placement}). A clear trend emerges: performance consistently improves as the module is moved deeper into the network, peaking at the Decoder - Penultimate layer. This position strikes an effective balance between the rich semantic context from deeper layers and the spatial fidelity from shallower ones. We further tested an extreme configuration by inserting LFM modules into all 16 main U-Net blocks. Despite a massive computational overhead (+12.80 TFLOPs), this approach failed to outperform the optimal single-module placement. These results decisively validate our choice of a single, efficient LFM module.
\begin{table}[h!]
  \centering
  \small
\caption{Ablation on the placement of the LFM.}
  \label{tab:lfm_placement}
  \resizebox{\columnwidth}{!}{%
  \begin{tabular}{@{}l cc cc c@{}} 
    \toprule
    \multirow{2}{*}{LFM Insertion Layer} & \multicolumn{2}{c}{NYUv2} & \multicolumn{2}{c}{KITTI} & \multirow{2}{*}{TFLOPs} \\
    \cmidrule(lr){2-3} \cmidrule(lr){4-5}
    & AbsRel$\downarrow$ & $\delta_1\uparrow$ & AbsRel$\downarrow$ & $\delta_1\uparrow$ & \\
    \midrule
    Baseline (no LFM) & 6.2 & 94.8 & 10.9 & 89.8 & -- \\
    \midrule
    Encoder - Early & 6.1 & 95.1 & 10.7 & 90.2 & +0.85 \\ 
    Middle Block & 5.6 & 96.2 & 10.4 & 91.3 & +0.85 \\
    Decoder - Early & 5.4 & 96.7 & 10.0 & 91.4 & +0.85 \\
    \textbf{Decoder - Penultimate} & \textbf{5.3} & \textbf{96.9} & \textbf{9.8} & \textbf{91.5} & \textbf{+0.85} \\
    Decoder - Final & 5.5 & 96.5 & 10.2 & 91.3 & +0.85 \\
    \midrule
    All Blocks & 5.3 & 96.9 & 9.9 & 91.5 & +13.60 \\
    \bottomrule
  \end{tabular}%
  }
\end{table}

\textbf{Ablation on the Number of Basis Filters.}
We evaluate the impact of the number of learnable spectral masks, $N$, on the LFM module's performance. As shown in Table~\ref{tab:ablation_n_experts}, expanding the filter bank from a single mask ($N=1$) to $N=4$ yields consistent performance gains, demonstrating that a more diverse basis set of frequency responses provides richer expressive capacity. However, further increasing the number to $N=8$ provides no additional performance improvement, despite doubling the parameter cost. Therefore, we select $N=4$ as our default configuration, as it offers the best trade-off between performance and model complexity.
\begin{table}[h]
  \centering
\caption{Ablation on the number of basis filters ($N$). This study evaluates the trade-off between performance, parameter cost, and computational cost (TFLOPs).}
  \label{tab:ablation_n_experts}
  \begin{tabular}{@{}lcccc@{}}
    \toprule
    Number of Filters ($N$) & NYUv2 & KITTI & Added Params & TFLOPs \\
    \midrule
    Baseline (no LFM) & 6.2 & 10.9 & -- & 45.00 \\
    \midrule
    1 (Single Filter) & 6.1 & 10.4 & +0.25M & 45.20 \\
    2 & 5.6 & 10.1 & +0.45M & 45.40 \\
    \textbf{4 (Ours)} & \textbf{5.3} & \textbf{9.8} & \textbf{+0.85M} & \textbf{45.80} \\
    8 & 5.3 & 10.0 & +1.65M & 46.60 \\
    \bottomrule
  \end{tabular}
\end{table}

\textbf{Ablation on LFM Design.}
We validate our LFM design through a sequential build-up analysis in Table~\ref{tab:lfm_design_ablation}. Starting from a baseline without frequency modulation, simply enabling a parameter-free FFT branch with a fixed identity mask already provides a performance boost. To enable content-aware filtering, we introduce the routing network. A minimal design using only a $1 \times 1$ convolution (PM) performs poorly, but performance significantly improves when prepended with a local feature extractor (LE). To further expand contextual awareness, we compare two non-local aggregation approaches: a large kernel convolution (LKC) and a self-attention (SA) layer. While LKC provides a moderate boost, the SA layer achieves the best performance, demonstrating the critical value of dynamically aggregating global context for spatially-variant routing. The full pipeline confirms that a learnable, context-aware filter is significantly more effective than a fixed or overly simplistic one.
\begin{table}[h!]
  \centering
  \small 
\caption{
    Ablation on the LFM design, validating the sequential build-up of the routing network ($\Phi_{\text{router}}$). 
    LE: Local Extractor, LKC: Large Kernel Conv., SA: Self-Attention, PM: Projection Mapping.
}
  \label{tab:lfm_design_ablation}
  \begin{tabular}{cccc|cc cc}
    \toprule
    \multicolumn{4}{c|}{Routing Network} & \multicolumn{2}{c}{NYUv2} & \multicolumn{2}{c}{KITTI} \\
    \cmidrule(lr){1-4} \cmidrule(lr){5-6} \cmidrule(lr){7-8}
    LE & LKC & SA & PM & AbsRel$\downarrow$ & $\delta_1$$\uparrow$ & AbsRel$\downarrow$ & $\delta_1$$\uparrow$ \\ 
    \midrule
    \multicolumn{4}{c|}{Baseline (No LFM)} & 6.2 & 94.8 & 10.9 & 89.8 \\
    \multicolumn{4}{c|}{LFM (Fixed Mask)} & 5.8 & 96.1 & 10.4 & 90.9 \\
    \midrule
    \ding{55} & \ding{55} & \ding{55} & \ding{51} & 5.9 & 95.7 & 10.5 & 90.6 \\
    \ding{51} & \ding{55} & \ding{55} & \ding{51} & 5.8 & 96.1 & 10.2 & 91.0 \\
    \ding{51} & \ding{51} & \ding{55} & \ding{51} & 5.6 & 96.4 & 10.0 & 91.2 \\
    \textbf{\ding{51}} & \ding{55} & \textbf{\ding{51}} & \textbf{\ding{51}} & \textbf{5.3} & \textbf{96.9} & \textbf{9.8} & \textbf{91.5} \\
    \bottomrule
  \end{tabular}
\end{table}

\textbf{Overall Component Contributions.}
As shown in Table~\ref{tab:main_component_ablation}, adding either LFM or BiasMap to the baseline model yields significant performance gains on both NYUv2 and KITTI. When combined, the two modules work effectively in concert, with the full VistaDepth model achieving the best overall performance. This demonstrates that enhancing frequency-domain awareness and rebalancing loss supervision are complementary and highly effective strategies for improving diffusion-based depth estimation.
\begin{table}[h!]
  \centering
  \small
  \setlength{\tabcolsep}{8pt}  
  \caption{Overall component contributions. Checkmarks (\ding{51}) and crosses (\ding{55}) denote active and inactive components, respectively.}
  \label{tab:main_component_ablation}
  
  \begin{tabular}{cc cc cc}
    \toprule
    \multicolumn{2}{c}{Components} &
    \multicolumn{2}{c}{NYUv2} &
    \multicolumn{2}{c}{KITTI} \\
    \cmidrule(lr){1-2} \cmidrule(lr){3-4} \cmidrule(lr){5-6}
    LFM & BiasMap & AbsRel$\downarrow$ & $\delta_1$$\uparrow$ & AbsRel$\downarrow$ & $\delta_1$$\uparrow$ \\ 
    \midrule
    
    \ding{55} & \ding{55}  & 6.2 & 94.8 & 10.9 & 89.8 \\
    \midrule
    
    \ding{55} & \ding{51} & 5.4 & 96.7 & 9.9  & 91.4 \\
    \ding{51} & \ding{55} & 5.3 & 96.9 & 9.8  & 91.5 \\
    \midrule
    
    \textbf{\ding{51}} & \textbf{\ding{51}} & \textbf{5.2} & \textbf{97.4} & \textbf{9.7} & \textbf{91.8} \\
    \bottomrule
  \end{tabular}
\end{table}
\section{Conclusion}
\label{sec:conclusion}

We introduced VistaDepth, a diffusion-based framework that excels at far-range MDE by addressing feature degradation and supervision bias. The LFM module enhances latent features via frequency-domain modulation, while the BiasMap mechanism adaptively reweights the training loss. Together, they enable VistaDepth to set a new state-of-the-art for diffusion-based MDE. Despite its strong performance, future work could accelerate the iterative diffusion process via knowledge distillation and explore data-centric methods to create more robust training sets for far-range scenes.

\bibliographystyle{IEEEtran}
\bibliography{sections/references}

\end{document}